%% file: paper.tex
\documentclass[a4paper,twoside]{article}

\usepackage{epsfig}
\usepackage{subcaption}
\usepackage{calc}
\usepackage{amssymb}
\usepackage{amstext}
\usepackage{amsmath}
\usepackage{amsthm}
\usepackage{multicol}
\usepackage{pslatex}
\usepackage{apalike}
\usepackage{SCITEPRESS}     

\usepackage{amssymb}
\usepackage{pdfrender}
\newcommand*{\boldcheckmark}{%
  \textpdfrender{
    TextRenderingMode=FillStroke,
    LineWidth=.5pt, 
  }{\checkmark}%
}

\usepackage{booktabs}
\usepackage{graphics}
\input{comandosnuevos}

\begin{document}

\title{Single-view 3D Body and Cloth Reconstruction under Complex Poses}

\author{\authorname{Nicolas Ugrinovic\sup{1}\orcidAuthor{0000-0002-1823-3780}, 
Albert Pumarola\sup{1}\orcidAuthor{ 0000-0003-4185-6991 }, 
Alberto Sanfeliu\sup{1}\orcidAuthor{0000-0003-3868-9678} and \\
Francesc Moreno-Noguer\sup{1}\orcidAuthor{0000-0002-8640-684X}
}
\affiliation{\sup{1}Institut de Robòtica i Informàtica Industrial, CSIC-UPC, Barcelona, Spain}
\email{\{nugrinovic, apumarolara, asanfeliu, fmoreno\}@iri.upc.edu}
}


\input{00_abstract}

\onecolumn \maketitle \normalsize \setcounter{footnote}{0} \vfill

\input{01_introduction}

\input{02_related}

\input{03_method}

\input{04_implementation}
\input{05_experiments}

\input{06_conclusions}

\bibliographystyle{apalike}
{\small
\bibliography{references}}

\end{document}

%% file: comandosnuevos.tex
\newcommand{\comment}[1]{}

\newcommand{\bd}{\mathbf{d}}

\newcommand{\bv}{\mathbf{v}}

\newcommand{\by}{\mathbf{y}}

\newcommand{\bI}{\mathbf{I}}

\newcommand{\bM}{\mathbf{M}}



%% file: 00_abstract.tex
\abstract{Recent advances in 3D human shape reconstruction from single images have shown impressive results, leveraging on deep networks that model the so-called implicit function to learn the occupancy status of arbitrarily dense 3D points in space. However, while current algorithms based on this paradigm, like PiFuHD~\cite{saito2020pifuhd}, are able to estimate accurate geometry of the human shape and clothes, they require high-resolution input images and are not able to capture complex body poses. Most training and evaluation is performed on 1k-resolution images of humans standing in front of the camera under neutral body poses. In this paper, we leverage publicly available data to extend existing implicit function-based models to deal with images of humans that can have arbitrary poses and self-occluded limbs. We argue that the representation power of the implicit function is not sufficient to simultaneously model details of the geometry and of the body pose. We, therefore, propose a coarse-to-fine approach in which we first learn an implicit function that maps the input image to a 3D body shape with a low level of detail, but which correctly fits the underlying human pose, despite its complexity. We then learn a displacement map, conditioned on the smoothed surface and on the input image, which encodes the high-frequency details of the clothes and body. In the experimental section, we show that this coarse-to-fine strategy represents a very good trade-off between shape detail and pose correctness, comparing favorably to the most recent state-of-the-art approaches. Our code will be made publicly available.}

%% file: 01_introduction.tex
\section{\uppercase{Introduction}} \label{sec:introduction}

While the 3D reconstruction of the human pose~\cite{Martinez_2017_ICCV,MorenoCVPR2017,PavlakosZDD17,RogezWS18,Mehta_3DV2018,kokkinos_3dpose_2018} and shape of the naked body~\cite{kanazawa_end--end_2017,pavlakos_learning_2018,varol2018bodynet,varol17_surreal} from single images has been extensively studied over the past few years and led to very accurate results, doing this with clothed humans remains a difficult challenge. There exist recent works that provide very good body and cloth reconstructions,  but are  methods limited to mild human poses, typically standing up in front of the camera~\cite{saito2020pifuhd,saito2019pifu,siclope,tex2shape,jackson_3d_2018}. A challenge that still remains open is thus to capture diverse poses while maintaining a detailed geometry of clothes and body. 

PiFu~\cite{saito2019pifu} and very recently~\cite{saito2020pifuhd} are the most relevant works on clothed human reconstruction, and builds upon the representation capacity of implicit functions, shown to be very effective for estimating the geometry of rigid 3D objects~\cite{mescheder_occupancy_nodate,im-gan,xu_disn:_2019}. PiFu learns 
a per-pixel feature vector aligned with the 3D surface to get an implicit function based on local information. However, while this strategy provides  a lot of detail, it cannot generalize to arbitrary human poses. 

Other works are able to capture diverse poses but lack details of human clothing~\cite{genova2019deep}. There exist methods that do not use implicit functions, but introduce an additional step to the estimation of a parametric naked body model. For instance,~\cite{tex2shape} learns a displacement map over the SMPL model~\cite{loper_smpl:_2015}, although, this approach is also limited to a small range of body poses and it needs high-quality $1024 \times 1024$ input images.

In this paper, we use implicit functions and propose an approach that, given a single image, is able to predict detailed meshes of clothed 3D humans for a wide range of poses and can work with but it is not limited to $224 \times 224$ input images. 

We argue that one of the reasons why~\cite{saito2019pifu,saito2020pifuhd} does not generalize well to difficult poses is that it strongly relies on local pixel features to guide the reconstruction and, thus, has no awareness of the overall topology of the mesh and therefore struggle to model unseen parts of the body. 

\begin{figure*}
\begin{center}
\includegraphics[width=\textwidth]{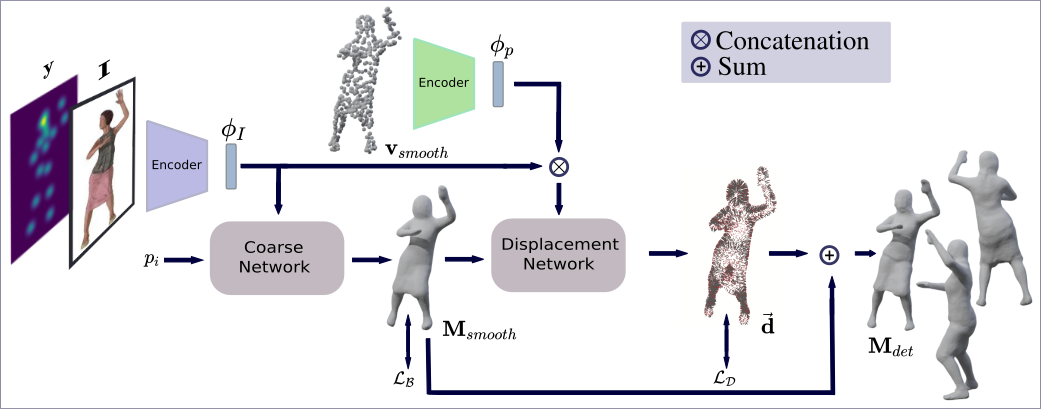}
\end{center}
\vspace{-3mm}
   \caption{{\bf Overview of our pipeline to reconstruct clothed people under complex poses.} Given an input RGB image, an  implicit function-based network initially predicts  a smoothed version of the geometry, but with an accurate body pose. The fine details of the mesh are recovered by a second network that computes a displacement field over the smooth mesh.} 
\label{fig:intro}
\vspace{-2mm}
\end{figure*}

To address this, we exploit global image features and alleviate their inherent lack in details using two strategies: First, we introduce a coarse-to-fine architecture with two modules, one building on an implicit function and global features that learns a  coarse 3D shape, but with a correct body pose; and another network that learns a displacement map to add extra detail (see Fig.~\ref{fig:intro}). Second, we take into account the  structure of the human body by including 2D joints as inputs of our system. This enables to have overall mesh consistency and retain the details of body and clothing in complex poses.

We quantitatively evaluate our method on synthetic data and qualitatively on real and synthetic images and demonstrate that our approach can capture a wide range of poses better than previous state-of-the-art methods based on implicit functions. Thus, we claim that global reasoning combined with a refinement step leads to coherent human meshes with no disconnected body parts, even in difficult poses, while maintaining a good level of detail.

%% file: 02_related.tex
\section{\uppercase{Related Work}} \label{sec:related_work}
\vspace{2mm}
\noindent{\bf Single-view 3D Reconstruction of Rigid Objects}
 is a well studied topic in computer vision and computer graphics. The works in this realm can be mainly categorized by the representation they use, whether it is a voxel grid~\cite{choy_3d-r2n2:_2016,voxels:TulsianiZEM17,voxels:MarrNet}, pointcloud~\cite{pumarola2019c,fan_point_2016}, mesh~\cite{wang_pixel2mesh:_2018,mesh-rcnn} or implicit function~\cite{mescheder_occupancy_nodate}. Voxels usually require extensive memory and are time consuming to train while usually leading to reconstructions with very restricted resolution. Pointclouds require additional non-trivial post processing steps to generate the final mesh. \cite{wang_pixel2mesh:_2018,mesh-rcnn} directly work on the mesh using a graph based CNN~\cite{graph-cnn}, although they are only able to generate overly smoothed meshes with  simple  topology which can be genus-0 only. In contrast, we choose to work with implicit function representation due to the well known fact that they require relatively simple architectures and have the ability to obtain a greater level of detail without requiring vast amounts of memory. 

Several works~\cite{mescheder_occupancy_nodate,park_deepsdf:_2019,xu_disn:_2019,im-gan} have shown that  implicit functions can be learned by means of deep neural networks, and it is  possible to get high resolution reconstruction by applying the marching cubes (MC) algorithm. Most recent approaches for image 3D reconstruction use implicit functions. For example,~\cite{mescheder_occupancy_nodate} conditions the learning of occupancy probabilities to an input image,  being able to reconstruct a high resolution mesh. However, they rely solely on global image features which hinders the model to learn high frequency details. We, instead, use local information about the joints and learn a displacement map to improve the reconstruction details as a result of the MC algorithm.~\cite{im-gan} also uses global features suffering from the same lack of detail needed to capture clothed humans. 

\vspace{2mm}
\noindent{\bf Single-View 3D Human Reconstruction.}
While the problem of localizing the 3D position of the joints from a single image has been extensively studied~\cite{Martinez_2017_ICCV,MorenoCVPR2017,RogezWS18,Moon_2019_ICCV_3DMPPE,Mehta_3DV2018} 3D human body shape reconstruction still remains an open problem. Single-view human reconstruction requires strong priors due to the inherent ambiguity of the problem. This has been addressed by using parametric models learned  from  body  scan  repositories such as SCAPE~\cite{Anguelov05scape:shape} and SMPL~\cite{loper_smpl:_2015} to represent the human body geometry by a reduced number of parameters. These parameters are then optimized to match image characteristics. For example, methods that use deep neural networks input additional information such as silhouettes~\cite{dibra17-sil,pavlakos_learning_2018} and other types of manual annotations~\cite{lassner_unite_2017,omran2018}. Furthermore,~\cite{Cipolla_BMVC2017_15} uses a differential renderer along with a deep neural network to predict SMPL body parameters by directly estimating and minimizing the error  of  image  features. Despite the usefulness of parametric models, they can only reproduce the geometry of the naked human body. 

Monocular reconstruction of cloth geometry has been traditionally addressed under the  Shape-from-Template (SfT) paradigm~\cite{Moreno_pami2013,Sanchez_cvpr2010,Moreno_cvpr2011b,Agudo_cviu2016}, requiring  3D-to-2D point correspondences between a template mesh and the input. More recently~\cite{Pumarola_cvpr2018b} introduced a deep network which alleviated the need for estimating correspondences. In any event, the clothes reconstructed by these approaches, were  focused to simple rectangular-like shapes, and were not applicable to reconstruct the shape of the garments worn by humans.

To overcome this limitation, ~\cite{tex2shape} proposes to learn a displacement map on top of the SMPL body model and is able to represent certain type of clothing, short hair details and hands details. However, it fails for more complex topologies such as dresses and skirts and it is limited to mild human body poses (people standing in front of the camera and looking at it). Also others use displacement maps for this purpose~\cite{zhu2019detailed,onizuka2020tetratsdf}, although mostly from videos or few image frames~\cite{alldieck2018video,alldieck2019learning}. In this paper, while we also learn a displacement map,  we are capable of capturing dresses and skirts while including a large diversity of body pose. 

To address the limitations of parametric models, template-free methods have been used, some based on voxel representations~\cite{varol2018bodynet,deephuman2019,jackson_3d_2018}, others based on different representations~\cite{pumarola_3dpeople:_2019,saito2019pifu}. BodyNet~\cite{varol2018bodynet} infers the volumetric body shape, although, due to resolution constrains and the use of SMPL as a final fitting, it cannot recover clothing geometry. DeepHuman~\cite{deephuman2019} uses a volume-to-volume translation approach showing impressive results to capture pose and certain type of clothing, but it fails to correctly capture complex cloth geometry such as skirts and also suffers from high memory requirements of voxel representation, limiting its resolution and requiring and initial estimation of template-based model SMPL. To tackle the resolution limitation of voxels, GimNet~\cite{pumarola_3dpeople:_2019} uses geometry images to represent the body shape and is able to capture complex poses and geometries such as dresses, although with a lack of details. Finally, PIFu~\cite{saito2019pifu} and PIFuHD~\cite{saito2020pifuhd} use implicit function representation which is memory efficient and results in impressive level of details even for complex cloth geometries and accessories. However, this approach can not generalize to arbitrary human poses. We also use an implicit function representation, but in contrast to previous approaches we are able to capture a large range of arbitrary poses. This is made possible thanks to  a first module of our model, which is general enough and reasons in a global manner generating realistic human meshes.

Finally, most similar in concept but very different implementation from our work, ~\cite{he2020geopifu} demonstrate that in order to have a better detailed reconstruction, it is first necessary to have a solid geometric prior which can be learned from a coarse voxel representation of the human body.

\vspace{2mm}
\noindent{\bf 3D datasets.}
Even though 3D reconstruction has become a popular topic in the field, there are very few publicly available datasets that contain 3D information of human body. Obtaining the 3D body shape is a complex task that requires vast amounts of effort. BUFF dataset~\cite{shape_under_cloth:CVPR17_buff} is one of the few that contains high-quality 3D scans, nevertheless, it only includes 6 different subjects and although it has a good human body pose variation, only captures restricted actions. As an alternative, datasets with synthetically photo-realistic images have appeared in the scene~\cite{varol17_surreal,pumarola_3dpeople:_2019}. SURREAL~\cite{varol17_surreal} is the largest dataset, containing 6 million frames generated by projecting synthetic textures of clothes onto random SMPL body shapes. However, given that clothes are projected onto a naked body model, they are only textures and have no shape of their own, making it impossible to learn clothing details from this dataset. On the contrary, 3DPeople~\cite{pumarola_3dpeople:_2019} contains models of 80 different 3D dressed subjects that perform 70 actions and 2.5 million photorealistic rendered images in which every action sequence is captured by from 4 camera views. For this work we use 3DPeople dataset. Most recently ~\cite{Caliskan_2020_ACCV} announced a similar dataset containing images of synthetic humans and their corresponding 3D human mesh annotations. We don't use this dataset, however, because it has not yet been made public.

%% file: 03_method.tex
\section{\uppercase{Method}} \label{sec:method}
\subsection{Problem Formulation}
We aim to solve the problem of single image 3D reconstruction applied to human bodies with clothing. Our goal is to  make sure that not only the inferred pose of the mesh representing the person is correct but also that we recover geometry details of the clothing. 

Let   $ {\bf I} \in \mathbb{R}^{H \times W \times 3} $ be an input RGB image of a single clothed person at an arbitrary pose.
Our aim is to learn a  mapping $\mathcal{M}$  to reconstruct the mesh \textbf{M} which is a detailed 3D representation of the clothed body of the person. We represent  \textbf{M} as a mesh with $N$ vertices $v_i$, where $v_i = (x_i, y_i,z_i)$ are the 3D coordinates of each vertex  that explains the body of the person in the image, taking into account the body shape, pose and clothing details. We train $\mathcal{M}$ in a supervised manner. 

\subsection{Network Architecture}
We next describe our network to generate detailed meshes under complex poses from a single image.
Given the high complexity of the task, we use a coarse-to-fine approach and divide our method into two main modules, as shown in Fig.~\ref{fig:intro}.  

The first module, denoted \textit{coarse network}, outputs a smoothed mesh $\textbf{M}_{smooth}$ provided an input set of query points $\textbf{p}$ and an observation of the 3D object, the image $ \textbf{I}$. This mesh {\em intentionally} lacks the level of detail we are looking for but it is enforced to accurately fit the body pose.

The second module, which we call \textit{displacement network},  adds details to the mesh by  estimating vertex displacement $\vec{d}_i$ over the direction of the normal vector $\vec{n}_i$ for each vertex $v_i$ of $\textbf{M}_{smooth}$, yielding to $\textbf{M}_{det}$. For this, we learn a   network that takes as inputs $\textbf{I}$ and a set of  vertices randomly sampled from  $\textbf{M}_{smooth}$, which we shall denote  $\textbf{v}_{smooth}$.  

It is worth noting that, as an additional input to guide the learning of both networks, we use the 2D joints of the person in \textbf{I}. Next, we explain both networks in  detail. 

\subsubsection{Coarse Network}

Given the input image \textbf{I}, we use 
$J$ ground truth 2D body joint locations and represent them as heatmaps $\by\in \mathbb{R}^{H \times W \times J}$. 
We use $J=17$ body joints. This joint representation is then concatenated with \textbf{I} and fed into the network. Additionally, the network has as input a set of query points in the 3D space $\textbf{p}_{xyz}=\{p_i\}_{i=1}^K$. Our goal is to learn the occupancy probability for each $p_{i}$
given \textbf{I} and \textbf{y}. Formally, we seek to estimate the mapping: 
\begin{equation}
\mathcal{M} : \bI \oplus \by, p_i \rightarrow [0, 1]\;
\end{equation}

This mapping takes the form of an implicit function and can be learned by a neural network  $\textit{f}_{\theta_{s}}(p_i, \bI, \by) $. Estimating  $\mathcal{M}$ to  account for high frequency details is, however, significantly challenging for the network, and indeed we found out that training this network to learn details resulted in meshes with incorrect body poses. For this reason, we force it to learn a smoothed version of the occupancy field of the ground truth mesh, hence its name \textit{coarse network}. To enforce this, instead of using the detailed mesh as ground truth, we train this network with a pseudo ground truth that results from applying Laplacian smoothing~\cite{laplacian}.

Finally, at inference, to recover the mesh we first evaluate    $\textit{f}_{\theta_{s}}(p, \bI, \by) $ for all $p$ of a discretized  volumetric space. We then use an octree based algorithm MISE~\cite{mescheder_occupancy_nodate} and mark each  $p$ as occupied if  $\textit{f}_{\theta_{s}}(p, \bI, \by) $ is bigger or equal than some threshold $\tau$.
After the evaluation is complete, we apply the MC algorithm~\cite{marching_cubes} to extract and approximate isosurface and estimate the faces topology of $\bM_{smooth}$. Note that although we intentionally train $\textit{f}_{\theta_{s}}$ to produce a smooth mesh, the body pose is expected to be correct. Also note that we build on~\cite{mescheder_occupancy_nodate} and, therefore, follow their formulation, however, any other reconstruction model could be used instead.

\subsubsection{Displacement Network}
This network has a similar architecture as the previous one with two main differences: instead of estimating occupancy probability, it regresses the magnitude for displacements $\vec{d}_i$  and takes an additional conditioning value that also serves as a query input, the vertices $\bv_{smooth}$. 
In the same fashion as before, we learn a new encoding for the image and joints representation $\phi_{I}$ but we use $\bv_{smooth}$ to generate a point encoding $\phi_{p}$ and concatenate this to $\phi_{I}$. This way we are able to condition the learning of the displacements on $\bI$, $\by$ and $\bM_{smooth}$. This network, denoted  $\textit{h}_{\theta_{d}}(p, \bI, \by, \bv) $,  
regresses the magnitude of the displacement $\vec{d}_i$ which is then applied to $\bv_{smooth}$ in the direction of the normals $\vec{n}_i$ of $\bM_{smooth}$. This reduces the complexity of the problem by forcing the regressor to learn only a scalar value and not a 3-dimensional vector, helping the network to learn the proper displacements. 

The final result is obtained by adding the learned displacements to the vertices estimated by the first module: 
\begin{equation}
\bv_{det} = \bv_{smooth} + \vec{\bd}\;,
\label{eq:2}
\end{equation}
where $\bv_{det}$ are the vertices that correspond to $ \bM_{det} $ and share the same faces as $\bM_{smooth}$ and $\vec{\bd}$ is the estimated displacement over the direction of the normal vector.

Finally, at inference, to obtain the detailed mesh $\bM_{det}$ we first evaluate $h_{\theta_{d}}(p, \bI, \by, \bv)$, that in this case are all vertices $\bv_{smooth}$.  Then, using equation~\ref{eq:2} we get the detail vertices for $\bM_{det}$.

\subsection{Learning the Model}

\subsubsection{Smooth Reconstruction Loss}
To learn the parameters $\theta_s$ of the neural network $\textit{f}_{\theta_s}(p, \bI, \by)$, 
we randomly sample points in the 3D bounding volume of the mesh representing the person. We sample these points in three ways: \textit{(a)} uniformly over the bounding volume, \textit{(b)} densely over the face and hands, and \textit{(c)} densely over the surface. For \textit{b} and \textit{c} we sample several points (much more than \textit{a}) near the surface of the mesh, that is why we say it is a dense sampling. To automatically obtain sampling points for face and hands we only sample points within a radius \textit{r} of a sphere centered at the 3D joints corresponding to hands and face. We found that hands and face require higher level of detail to be better reconstructed than feet, hence, we do not include sampling specifically corresponding to feet.
For each sample image $i$ in  a training batch we sample \textit{K} points $\textit{p}_{ij} \in \mathbb{R}^3, j=1, ..., K$. The minibatch loss $\mathcal{L}_{B}$  is then is evaluated at those locations: 
\begin{equation}
\mathcal{L}_{\mathcal{B}}(\theta_{s}) = \frac{1}{B} \sum_{i=1}^{|\mathcal{B}|} \sum_{j=1}^{K}\mathcal{L}(f_{\theta_{s}}(p_{ij}, \bI, \by),o_{ij})\;,
\end{equation}
where $o_{ij} \equiv o(p_{ij})$ denotes the true occupancy at point $p_{ij}$, and $|\mathcal{B}|$ is the minibatch size. The loss $\mathcal{L}(\cdot, \cdot)$, different from~\cite{mescheder_occupancy_nodate}, is a weighted binary cross-entropy (wBCE) classification loss that takes into account the unbalanced number of points that lay inside the mesh in contrast to those that are outside. This avoids losing important body parts, especially the limbs, when extracting the mesh. 

In a similar fashion as in~\cite{mescheder_occupancy_nodate} we also introduce a generative loss that helps us capture the rich distribution of complex clothing. We do this by adding an encoder network $g_{\psi}(\cdot)$ that takes as inputs the points and occupancies to predict the mean $u_{\psi}$ and standard deviation $\sigma_{\psi}$ of a Gaussian distribution $q_{\psi}(z|(p_{ij}, o_{ij})_{j=1:K})$ on a latent space $z \in \mathbb{R}^L$ as output and then optimizing the KL divergence. This way, the new loss becomes:

\begin{equation}
\begin{split}
\label{eq:4}
\mathcal{L}_{\mathcal{B}}(\theta) = &\frac{1}{B} \sum_{i=1}^{|\mathcal{B}|}[ \sum_{i=j}^{K}\mathcal{L}(f_{\theta_s}(p_{ij}, \bI, \by),o_{ij}) + \\[1ex]
&KL(q_{\psi}(z|(p_{ij}, o_{ij})_{j=1:K})||p_0(z))]\;
\end{split}
\end{equation}

\vspace{1mm}
where $p_0(z)$ is a prior distribution on the latent variable $z_i$ and $z_i$ is sampled according to $q_{\psi}(z|(p_{ij}, o_{ij})_{j=1:K})$.  We train this as a conditional variational autoencoder~\cite{cvae}.    

To generate $\bM_{smooth}$  we use a hierarchical iso-surface extraction algorithm~\cite{mescheder_occupancy_nodate}, that incrementally builds an octree to efficiently obtain a high resolution mesh, that is then forwarded to the second stage of our method.

\subsubsection{Displacement Loss}
In order to learn the parameters $\theta_{d}$ of the neural network $\textit{h}_{\theta_{d}}(p, \bI, \by, \bv)$, in a similar manner as we did  with the coarse network, we randomly sample \textit{N} points $p_{ij}$ from $\bv_{smooth}$ and evaluate the minibatch loss $\mathcal{L}_\mathcal{B}$. Yet,  instead of using a wCBE loss, we use an L2 loss:
\begin{equation}
\mathcal{L}_{\mathcal{D}}(\theta_{d}) = \frac{1}{B} \sum_{i=1}^{|\mathcal{B}|} \sum_{j=1}^{N}  \|f_{\theta_{d}}(p_{ij}, \bI, \by, \bv_{smooth}) - d_{ij}\|_2
\end{equation}

%% file: 04_implementation.tex
\section{\uppercase{Implementation Details}} \label{sec:implementation}

Our model builds upon the ONet network architecture ~\cite{mescheder_occupancy_nodate}. For the \textit{coarse network} $\textit{f}_{\theta_s}(p, \bI, \by)$ we use 5 ResNet blocks~\cite{cvpr2016:Resnet} which are conditioned on  the  input  using  conditional  batch  normalization~\cite{batchnorm}. For the image and joint encoding we use a ResNet18 architecture.

For the \textit{displacement network} we modify the architecture by adding 5 ResNet blocks (yielding to a total of 10 blocks) 
and changing the last layer to regress the displacement value. We observed that for less amount of layers, the network is not able to capture the complexities of clothes and other details. It is important for the network to understand the 3D structure of the body in order to regress the desired displacements, for this reason we also modify the conditioning input of the architecture to be able to include mesh vertices as a prior. For this we use a similar encoder as in PointNet~\cite{cvpr2017:PointNet} and for the network we use 10 ResNet blocks. We plan to release our code.

\begin{figure*}[t!]
\begin{center}
\includegraphics[width=0.99\linewidth]{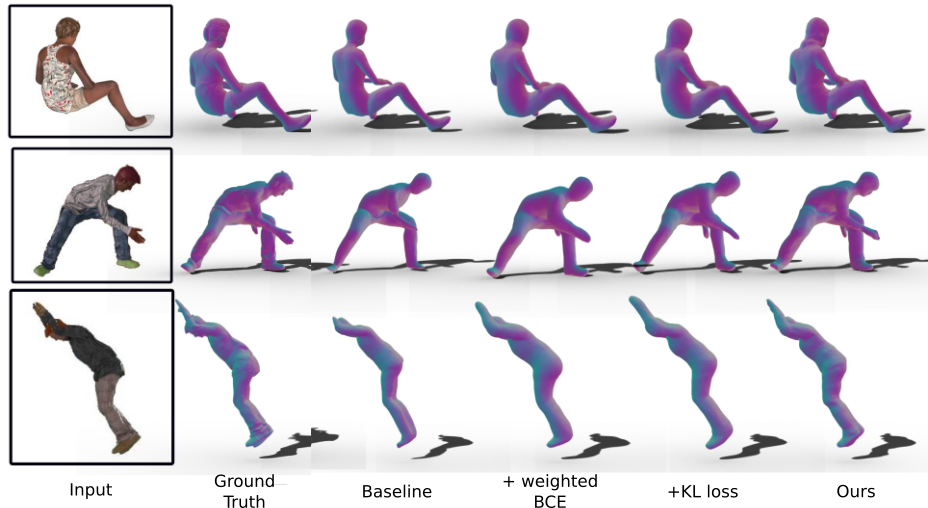}
\end{center}
\vspace{-1mm}
   \caption{{\bf Comparison between the baselines used on 3DPeople.} Baseline is~\cite{mescheder_occupancy_nodate} retrained on 3DPeople and subsequent columns are results of added components to that model validated in Table \ref{tab:ablation}. The figure displays the reconstructed meshes from the camera viewpoint. The color of the meshes encodes the normal directions of the surface. Note how our approach captures the global consistency of the mesh, as the previous, and additionally presents certain clothing details. }
   \vspace{1mm}
\label{fig:comparison}
\end{figure*}

The model is trained with 60,000 synthetic images of cropped clothed people resized to 224 x 224 pixels as needed by the image encoder, however, this resolution could be easily changed. These images correspond to  15,000 different meshes of varying number of vertices taken from the 3DPeople dataset~\cite{pumarola_3dpeople:_2019} and projected to 4 camera views. We use 44 subjects out of 80 to reduce training time.

\begin{table}
  \caption{ Quantitative evaluation on 3DPeople. Numerical comparison of our approach with other methods that use implicit functions retrained with the same data as ours. We measure IoU, Chamfer distance,  Normal Consistency  and Point to Surface  (see main text) to validate the different components of our model. $\uparrow$: higher the better. $\downarrow$: lower the better.}
  \label{tab:metric_results}
  \vspace{1mm}
  \resizebox{\columnwidth}{!}{%
  \begin{tabular}{l|c|c|c|c}
    \toprule
    Method & IoU $\uparrow$ & Chamfer $\downarrow$ &  {\begin{tabular}[c]{@{}c@{}}Normal\\ Consistency\end{tabular}}$\uparrow$& P2S $\downarrow$ \\
    \midrule
ONet &      0.516       &   0.280       &   0.793          & 18.135 \\
PiFu &      0.244       &   1.550       &   0.601          &70.200 \\
Ours & \textbf{0.610} & \textbf{0.100} & \textbf{0.821} & \textbf{16.200} \\
    \bottomrule
  \end{tabular}
  } %
\end{table}

\begin{table*}
  \caption{Quantitative ablation study on 3DPeople dataset. Note that dense sampling denotes sampling strongly in the surface of the mesh (meaning several more points than in uniform sampling)}
  \label{tab:ablation}
  \resizebox{\textwidth}{!}{%
    \begin{tabular}{ccccccc|c|c|c|c|c}

  \cmidrule(lr){1-7} \cmidrule(lr){8-11}
  \multicolumn{7}{c}{\textit{Components}} & \multicolumn{4}{c}{\textit{Metrics}}\\
  
    Occupancy & wBCE & KL & Joints & Uniform Samp. & Dense Samp.  & Displacement & CD $\downarrow$&IoU$\uparrow$&Normal Consistency$\uparrow$&P2S$\downarrow$\\
    \midrule
    \boldcheckmark&             &               &               &\boldcheckmark&             &                           &2.752 & 0.516 & 0.793 & 18.135\\
    \boldcheckmark&\boldcheckmark&              &               &\boldcheckmark&             &                            &1.689 & 0.576 & 0.808 & 18.698\\
    \boldcheckmark&\boldcheckmark&\boldcheckmark&               &\boldcheckmark&             &                            &1.496 & 0.579 & 0.811 & 18.353\\
    \boldcheckmark&\boldcheckmark&\boldcheckmark&               &\boldcheckmark&\boldcheckmark&                           &1.422 & 0.579 & 0.814 & 18.265\\
    \boldcheckmark&\boldcheckmark&\boldcheckmark&\boldcheckmark&\boldcheckmark&\boldcheckmark&                       &\textbf{1.051} & \textbf{0.612} &\textbf{0.829} & 16.397\\
    \boldcheckmark&\boldcheckmark&\boldcheckmark&\boldcheckmark&\boldcheckmark&\boldcheckmark&  \boldcheckmark      & 1.082& 0.606 & 0.821 & \textbf{16.200}\\
    
  \bottomrule
\end{tabular}
}%
\end{table*}

In order to train $\textit{f}_{\theta_s}$ we generate occupancy annotations, i.e determine which points lie in the interior of the mesh. This step requires a watertight mesh. 
To do this we use code provided by~\cite{mescheder_occupancy_nodate}. We train the \textit{coarse network} during 645 epochs, K=2048 and  Adam~\cite{adam} optimizer with initial learning rate of $1e-4$, beta1 0.9, beta2 0.999. For weighted-BCE we use a positive weight of 25. For reconstructing the mesh, we use a threshold parameter $\tau$=0.96 for all cases. For this network to better capture complex poses, we first normalize each mesh w.r.t. three points: hips, upper left leg and upper right leg. 

To train $\textit{h}_{\theta_{d}}(p, \bI, \by, \bv) $ we generate ground truth data using the results obtained from our \textit{coarse network} and compute the displacement over the normal by first densely sampling the surface of the ground truth mesh and then finding the distance over the normal direction from a mesh vertex to the nearest point in the ground truth mesh. 
We train during 1700 epochs with batch size 14, K=2,048 and N=10,000. As for the optimizer we use Adam~\cite{adam} with initial learning rate of $1e-4$, beta1 as 0.9, beta2 as 0.999. At epoch 170 we change the learning rate to $1e-5 $ and, again, at epoch 1,200 to $1e-6$.

\begin{figure}[t]
\begin{center}
   \includegraphics[width=0.99\linewidth]{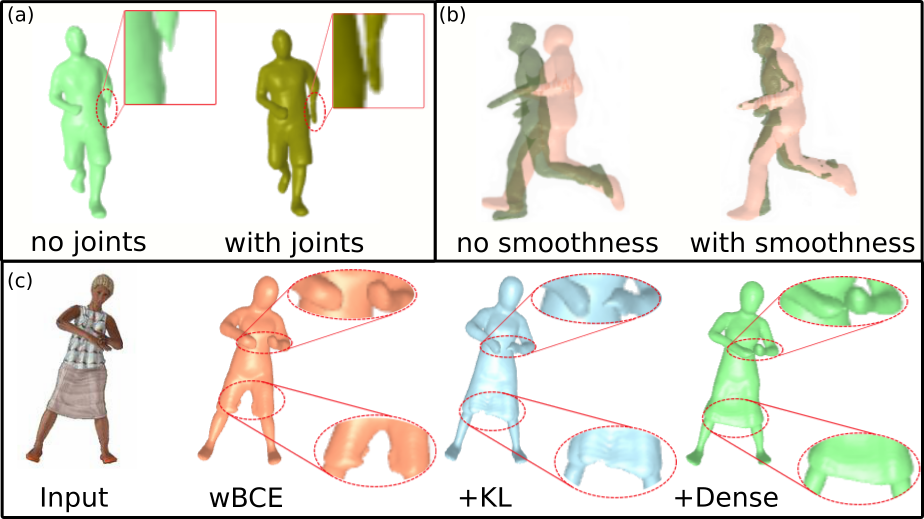}
\end{center}
\vspace{-1mm}
   \caption{\textbf{Visual ablation study.} Note that these are outputs of the \textit{coarse network} so they lack finer details. (a) Difference in reconstructions when using 2D joints as inputs vs. not using them.  (b) Effect of enforcing smoothness. Green meshes are ground truth, pink ones are reconstructions. Here we present reconstruction when the network tries to learn the detailed ground truth mesh vs. reconstruction when forcing \textit{coarse network} to learn a smooth version of the ground truth. (c) Impact of using a generative loss and "dense" sampling, meaning, sampling more heavily on the surface of the mesh. Adding generative loss helps to capture the richness of the 3D shape distribution. Here we can see that by adding KL loss and then sampling near the surface, especially around face and hands (dense), we obtain better results both in hands, face and skirts.(+KL=wBCE+KL, +Dense=wBCE+KL+Dense)}
\label{fig:ablation}
\vspace{1mm}
\end{figure}


%% file: 05_experiments.tex
\section{\uppercase{Experimental Evaluation}} \label{sec:experiments}

This  section  provides  an  evaluation  of  our  proposed method. We present quantitative and  qualitative  results on synthetic images from 3DPeople~\cite{pumarola_3dpeople:_2019} and qualitative results on images in the wild. We evaluate our approach on 3,200 images randomly chosen for 5 subjects (2 female/ 3 male) from~\cite{pumarola_3dpeople:_2019}.

\begin{figure*}[t!]
\begin{center}
\includegraphics[width=\textwidth]{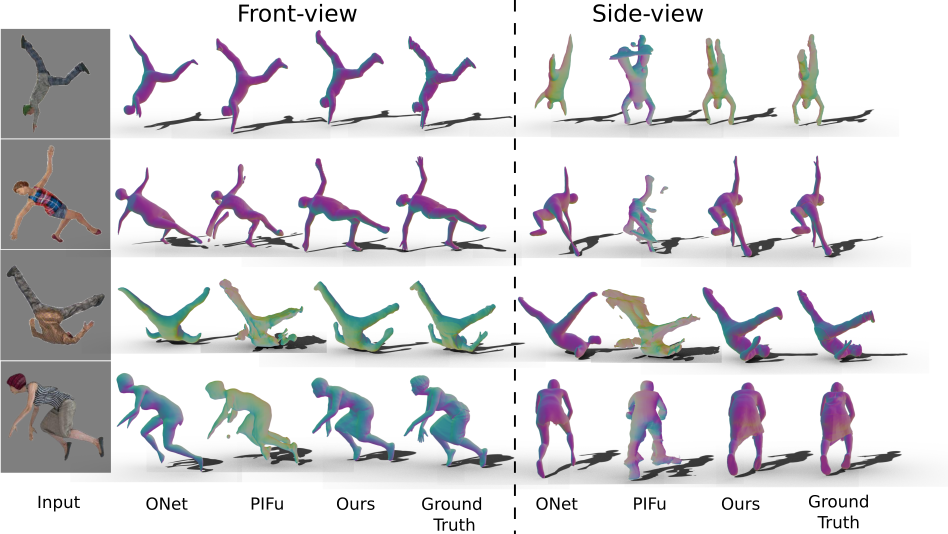}
\end{center}
\vspace{-1mm}
   \caption{{\bf Results on synthetic images of the 3DPeople dataset.} For every row we display the input RGB image and the mesh reconstructed using our approach and comparative approaches seen from two different viewpoints, Onets~\cite{mescheder_occupancy_nodate} and PIFu~\cite{saito2019pifu}.  The color of the meshes encodes the normal directions of the surface. }
\label{fig:synthetic_results}
\vspace{1mm}
\end{figure*}

We compare our approach quantitatively (see Table~\ref{tab:metric_results}) with other two prominent implicit function models for 3D reconstruction, namely, OccupancyNets (ONets)~\cite{mescheder_occupancy_nodate} and PiFu~\cite{saito2019pifu}. Note that to ensure fair comparison both a re-trained with the same training data as our model and we test all models with the same test set as ours. Although one could argue that numerical comparison with SOTA should include PIFuHD~\cite{saito2020pifuhd}, this was not possible as the authors have not released the training code. However, we believe that the methods in question are good representatives of powerful implicit function models for 3D reconstruction. In this sense, being ONet a good candidate for global consistency models and PIFu for hi-detail local consistent models. Qualitative comparison with both these methods on synthetic images can be found in Fig.~\ref{fig:synthetic_results}.

Additionally, in Table~\ref{tab:ablation} we present a quantitative ablation study to validate all the components propose in this paper and used by our final method. The table compares our method and several baselines built upon the Occupancy Net~\cite{mescheder_occupancy_nodate} and the losses we have defined in our system. Table~\ref{tab:ablation} reports  the errors for all methods and shows that  our approach consistently improves all baselines. Also, notice how the addition of the wBCE and KL losses over the ONet baseline, gracefully reduce the errors.

As evaluation metrics we use volumetric IoU, Chamfer distance (CD), normal consistency score and point to surface score (P2S). Volumetric IoU is defined as the quotient of the volume of the two meshes union and the volume of their intersection. We use the same procedure as in~\cite{mescheder_occupancy_nodate} to obtain this value. We calculate the CD by randomly sampling 100,000 points from both the watertight ground truth and the estimated meshes. We define a normal consistency score as the mean absolute dot product of the normals in one mesh and the normals at the corresponding nearest neighbors in the other mesh. 
As in~\cite{saito2019pifu},  we measure the average point-to-surface Euclidean distance (P2S) in cm from the vertices on the reconstructed surface to the ground truth.

Fig.~\ref{fig:comparison}  shows three samples of the meshes reconstructed with each of the baselines and our final method. Regarding the three ONet baselines, note how the introduction of the losses tend to produce better reconstructions, although the sharper geometry details are more evident in our approach (Fig.~\ref{fig:comparison}(ours)), which includes all previous losses plus the refinement of the geometry estimated with the displacement network. Also the effect of other components of our model and the proposed training scheme if depicted qualitatively in Fig.~\ref{fig:ablation}.

\begin{figure*}[t!]
\begin{center}
\includegraphics[width=\textwidth]{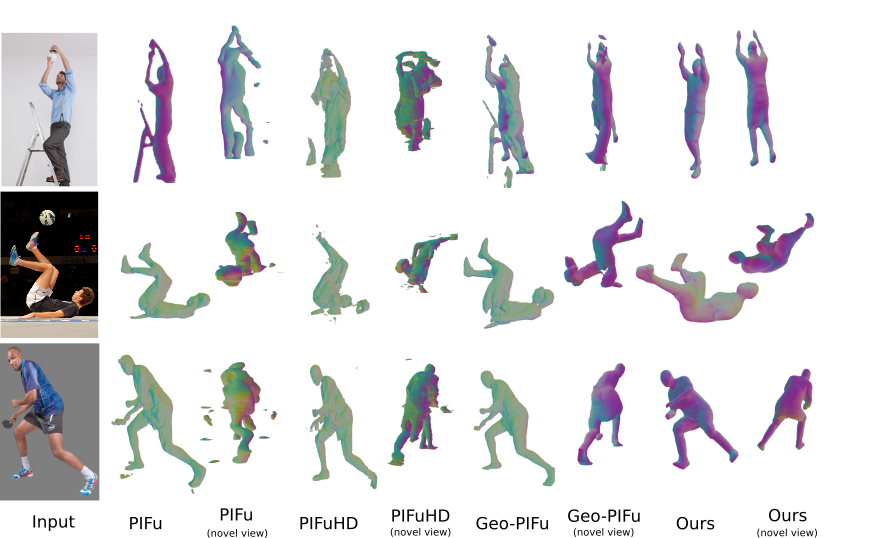}
\end{center}
\vspace{-1mm}
   \caption{{\bf Qualitative results of our approach on real images.} We compare with PIFu~\cite{saito2019pifu}, PIFuHD~\cite{saito2020pifuhd} and Geo-PIFu~\cite{he2020geopifu}}
\label{fig:real_results}
\vspace{-2mm}
\end{figure*}

Qualitative comparison on synthetic and real images is also presented. Fig.~\ref{fig:synthetic_results} presents sample synthetic images from our test set that none of the models have seen before. Here we present results of our method along with ONet and PIFu, note that all are re-trained with 3DPeople dataset. As shown in Fig.~\ref{fig:synthetic_results} one can note that PiFu if capable of reconstructing in a very acceptable manner all the front-view parts of the meshes, however, it fails to give a global consistency to the mesh. This can be seen in the columns depicting the side-view. We argue that this is due to PIFu's heavy reliance on local aligned features. Also we argue that PIFu is penalized by the relatively low-resolution of the input images, whereas our methods in not that sensitive to low-resolution failures. Additionally, since the 2D joints are not exploited by PiFu, the structure of the body it produces is not always consistent. Qualitative comparison with other SOTA methods on real images can be found in Fig.~\ref{fig:real_results}. Here we compare our method with ~\cite{saito2019pifu,saito2020pifuhd,he2020geopifu}. Note that non of these methods nor ours have been train with real images and inference, in this case, is done with the trained weights provided by the authors of each method. All PIFu methods, except Geo-PIFu (to a lesser extent) show the same problem addressed before: shockingly good front views, however lacking global consistency and human body coherence. Geo-PIFu works better in these cases as this model specifically aims for global coherence just as our method does.

\textbf{Impact of using 2D joints.} We found out that using 2D joints as additional input to our model improves the reconstruction quality. By adding joint information we prevent the network from generating incomplete human bodies, especially in cases where the image presents self-occlusions (see Fig.~\ref{fig:ablation}(a)).

\textbf{Impact of enforcing smoothness.} As stated before, we enforce the \textit{coarse network} to learn a smooth version of the ground truth mesh. This reliefs the network from learning a more complex mapping to account for high-level details which has an impact on the correctness of the reconstructed human pose. As shown in Fig.~\ref{fig:ablation}(b), one can clearly see that when we do not enforce to learn a smooth version of the mesh, the pose deviates considerably from the ground truth.

\textbf{Impact of using a generative model.} The use of generative loss (equation~\ref{eq:4}) helps the model to better capture the richness and variability of the distribution of human clothing and body details such as hands and face. As it can be seen in Fig.~\ref{fig:ablation}(c), when adding the KL loss term to the model the skirt and hands are better reconstructed. Moreover, this is improved when combining this with the dense sampling strategy that was mentioned before. 

\textbf{Impact of dense sampling.} When combined with the KL loss, the dense sampling strategy (near surface and around face and hands) helps the model to better capture the correct structure of clothing and human body. In the case of Fig.~\ref{fig:ablation}(c), we show how adding this sampling strategy results in better hands and skirts. Although not shown here, we also observed slight improvement in the face area.

\textbf{Real images.} We finally show in Figures~\ref{fig:synthetic_results} and~\ref{fig:real_results} the reconstructed shapes on synthetic and real images, respectively. Note, specially in the synthetic examples, how we are able to capture very complex body poses together with the details of the clothing (e.g. skirts). Also, note that for test and real images (given that we do not have ground truth for 2D joints) we use an off-the-shelf 2D pose detector such as~\cite{openpose}. Another alternative is to use~\cite{frankmocap} and get all necessary joints by projecting them into the 2D space.

Although we get good results on real images, it can be perceived, in some cases, that the results are not as good as on synthetic ones. We hypothesize that this is due to a slight difference in appearance of real images in contrast to synthetic ones, especially due to lighting conditions, shadows and color. It is known that there is domain gap between real and synthetic images. We believe that by training with real images or paying more attention to the photo-realism of synthetic images we would get even better results. While we are able to capture skirts, where most of other methods fail, there is still room for improvement. However, we believe that combining global reasoning with a refinement step to add details is the right direction to obtain coherent human meshes in a wide range of poses with high enough detail.

%% file: 06_conclusions.tex
\section{\uppercase{Conclusions}} \label{sec:conclusions}

In this paper we have made the following contributions to the problem of reconstructing the shape of dressed humans. As far as we can tell we are the first ones to do 3D reconstruction of clothed human body from single image in a wide range of poses including complex ones. In doing so, we do not require high resolution images. We demonstrate that different sampling schemes can improve the details with implicit function representation. 
Finally, we are able to capture details such as dresses and skirts while maintaining consistency of the body from all directions and not only the observed view.  

\section{Acknowledgments}
This work is supported  by the Spanish government with the projects MoHuCo PID2020-120049RB-I00 and María de Maeztu Seal of Excellence MDM-2016-0656.